\documentclass[runningheads]{llncs}
\usepackage{graphicx}
\usepackage{amsmath,amssymb} 
\usepackage{color}
\usepackage{multirow}
\usepackage{bbding}

\begin{document}

\title{Efficient Multi-level Correlating for Visual Tracking} 
\titlerunning{Efficient Multi-level Correlating for Visual Tracking} 


\author{Yipeng Ma\inst{1}
\and
Chun Yuan\inst{2}\inst{(}\Envelope\inst{)}
\and
Peng Gao\inst{1}
\and
Fei Wang\inst{1}\inst{(}\Envelope\inst{)}}
\renewcommand{\thefootnote}{}

%

\authorrunning{Y. Ma, et al.} 


\institute{Shenzhen Graduate School, Harbin Institute of Technology, China\\
 \and
Graduate School at Shenzhen, Tsinghua University, China\\
\email{\{mayipeng, pgao\}@stu.hit.edu.cn, wangfeiz@hit.edu.cn}\\
\email{yuanc@sz.tsinghua.edu.cn}
}

\maketitle


\begin{abstract}
Correlation filter~(CF) based tracking algorithms have demonstrated favorable performance recently. Nevertheless, the top performance trackers always employ complicated optimization methods which constraint their real-time applications. How to accelerate the tracking speed while retaining the tracking accuracy is a significant issue. In this paper, we propose a multi-level CF-based tracking approach named MLCFT which further explores the potential capacity of CF with two-stage detection: primal detection and oriented re-detection. The cascaded detection scheme is simple but competent to prevent model drift and accelerate the speed. An effective fusion method based on relative entropy is introduced to combine the complementary features extracted from deep and shallow layers of convolutional neural networks~(CNN). Moreover, a novel online model update strategy is utilized in our tracker, which enhances the tracking performance further. Experimental results demonstrate that our proposed approach outperforms the most state-of-the-art trackers while tracking at speed of exceeded 16 frames per second on challenging benchmarks.

\end{abstract}


\section{Introduction}
\label{sec:1}
Visual object tracking has made considerable progress in the last decades, and is widely developed in numerous applications, such as intelligent video surveillance, self-driving vehicle and human computer interaction. Despite the great effort that has been made to investigate effective approaches~\cite{ma2015long,valmadre2017,pgao180408208,pgao180407459,pgao2018ieice}, visual object tracking is still a tough task due to complicated factors like severe deformation, abrupt motion, illumination variation, background clutter, occlusion, etc. Due to requirement from many demanding applications, boosting both tracking speed and accuracy has long been pursued.

Recently, CF-based trackers have drawn considerable attention owing to their high tracking speed and good performance. Bolme et al.~\cite{bolme2010} are the first to exploit CF for visual tracking. Since then, several extended works are engaged in improving tracking performance. Henriques et al. propose CSK~\cite{henriques2012} and KCF~\cite{henriques2015} successively, which introduce circulant structure to interpret CF and generalize to the extension of multi-channel features. Additionally, Danelljan et al.~\cite{danelljan2014} exploit fast scale pyramid estimation to deal with scale variations. Despite the prominent efficiency of these CF-based trackers, intensive computation overhead due to the complex framework hinders its application in real-time scenarios.

To address the unwanted model drifting issue, Ma et al.~\cite{ma2015long} propose a complementary re-detection scheme based on an online random fern classifier. Also to address the same issue, Wang et al.~\cite{wang2017} conduct a multi-modal target re-detection technique with a support vector machine~(SVM) based tracking method. However, because of the directionless re-detection and too much proposals, inaccuracy and redundant computation do exist in these works.

With the great representation capability of deep features, convolutional neural network~(CNN)~\cite{lecun1998} has become popular in a wide range of computer vision tasks like object detection~\cite{girshick2014,sermanet2013} and object recognition~\cite{karpathy2014}. Most recently, CNN has been employed for visual tracking and shown its promising performance. Several CNN-based tracking approaches~\cite{bertinetto2016,qi2016,pgao2018} have shown state-of-the-art results on many object tracking benchmarks. These methods validate the strong capacity of CNN for target representation.

Inspired by previous works, we propose an efficient \emph{multi-level} CF-based framework for visual tracking. Here the so-called \emph{Multi-level} has two meanings: (a) Multiple layers of CNN are used to represent the target. Shallow and deep layers of CNN take the complementary properties into account. (b) A two-level detection scheme is proposed, i.e., primal detection and re-detection. Primal detection is cascaded with an oriented re-detection module. The primal detection delivers the possible candidate locations of the target to the re-detection module. Then, the re-detection module will conduct estimations around those locations and the most possible location is given as the location of the target finally.

The main contributions of our work can be summarized as follows:
    \noindent
    \begin{itemize}
    \item We propose a multi-level CF-based tracking method with features extracted from multiple layers of CNNs. Additionally, an effective fusion method based on relative entropy is applied to improve the tracking performance.
    \item We employ an oriented re-detection technique to ensure the localization accuracy. Furthermore, an effective adaptive online model update strategy is applied in our tracker.
    \item We compare our approach with state-of-the-art trackers on several benchmarks: OTB-2013, OTB-2015 and VOT-2017. The results validate that our tracker outperforms the most state-of-the-art trackers in terms of accuracy, robustness and speed.
    \end{itemize}

\section{Algorithmic Overview}
\label{sec:2}
\begin{figure*}[t]
        \begin{center}
        \begin{minipage}{\linewidth}\centerline{\includegraphics[width=\textwidth]{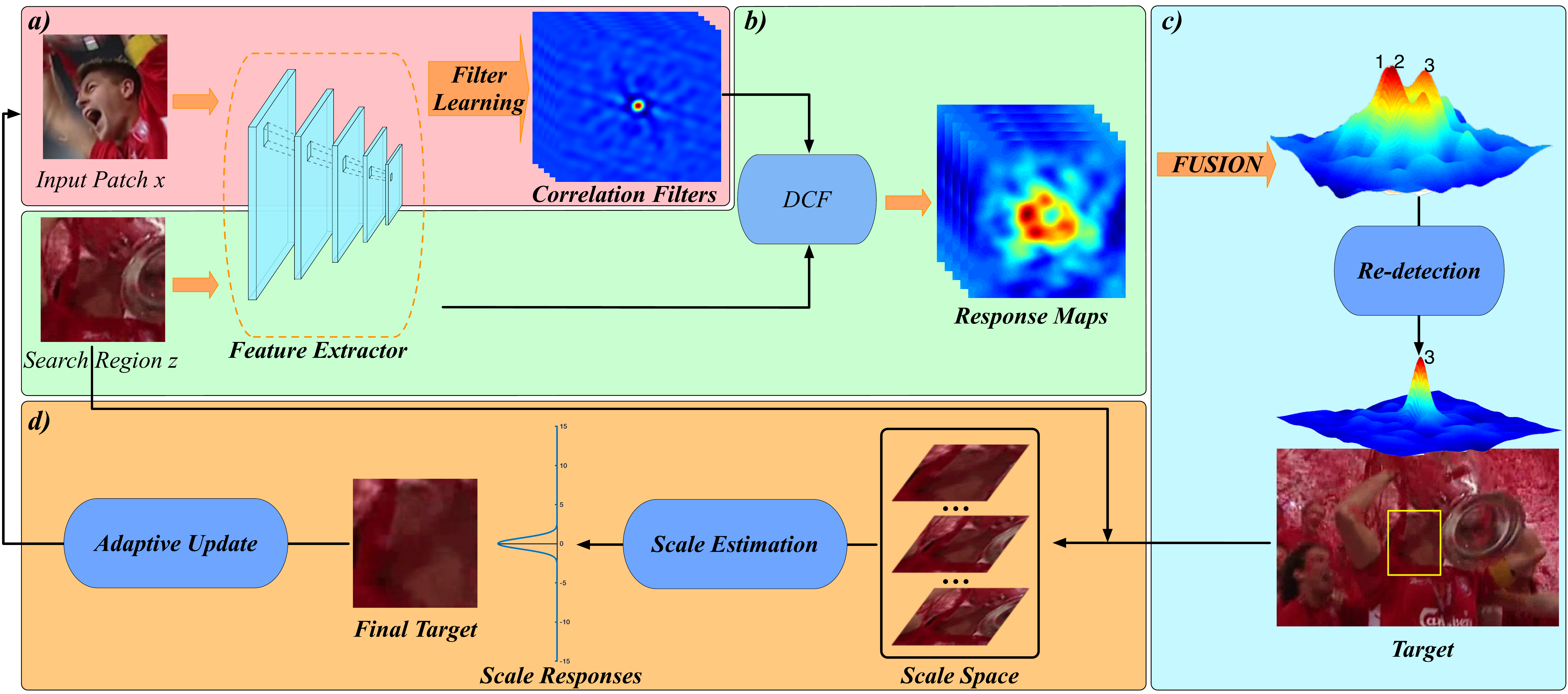}}\end{minipage}
        \end{center}
        \caption{The overall framework of the proposed method. The tracker can be divided into four parts marked in color: a) \textcolor[rgb]{1.00,0.00,0.00}{Filter learning}; b) \textcolor[rgb]{0.00,1.00,0.00}{Primal detection}; c) \textcolor[rgb]{0.00,0.07,1.00}{Re-detection module}; d) \textcolor[rgb]{1.00,0.50,0.00}{Scale estimation and adaptive update}.}
        \label{fig:1}
    \end{figure*}

A brief introduction to the overall framework of our proposed tracker is shown in Fig.~\ref{fig:1}. We divide our algorithm into four stages: filter learning, primal detection, re-detection and scale estimation followed by adaptive update. In the filter learning stage, we utilize the pre-trained VGG-Net to extract the feature maps of different convolutional layers from image patch. Then, the corresponding correlation filters are learned with these features and a Gaussian shaped label. In the primal detection stage, each feature map extracted from the search region is correlated by the corresponding correlation filter to generate response maps, respectively. And candidate locations of the target can be obtained in the fusion response map. In case of detection failure, the re-detection module is exploited in our tracker, which effectively avoid target drift during tracking (see section~\ref{subsec:32} for more details). In the last stage, scale estimation and a useful adaptive online model update strategy are applied to adjust scale variation and adapt model variation of the target.

\section{The Proposed Approach}
\label{sec:3}
In this section, we first describe the overall framework of our proposed approach in section~\ref{subsec:31}. Then we detail our proposed re-detection module in section~\ref{subsec:32}. Finally, we present the adaptive online model update scheme in section~\ref{subsec:34}.

\subsection{Multi-level Correlation Tracking}
\label{subsec:31}
Our tracking framework is constructed with the combination of canonical CF-based tracking approach with convolutional deep features The pre-trained VGG-Net~\cite{chatfield2014} is used to extract the convolutional features to represent the target. We observe the fact that the convolutional features extracted from very deep layer of CNN can capture rich discriminative semantic information, while the features from shallow layers offer more spatial information which is crucial to visual tracking. Therefore, multiple layers of VGG-Net are used to construct several weak trackers and then a fusion method based on Kullback-Leibler~(KL) divergence is proposed to unitize response maps produced by weak trackers to obtain an enhanced one.

CF models the appearance of a target using filters $w$ trained over samples $x$ and their corresponding regression target $y$. Given a feature map extracted from the $k$-th convolutional layer, denoted $x^k\in \mathbb{R}^{V\times H\times D}$, where $V$, $H$ and $D$ denote the height, width and the number of feature channels, respectively, and a Gaussian shaped label matrix $y\in \mathbb{R}^{V\times H}$ indicates the regression target. Then, the desired corresponding filter of the $k$-th convolutional layer can be obtained by minimizing the output ridge loss in the Fourier domain:
    \begin{equation} \label{eq:1}
        \hat{w}^k=\|\sum_{d=1}^D\hat{w}_d\scriptstyle\bigodot\textstyle\hat{x}_d^k-\hat{y}\|^2_F+\lambda\|\hat{w}\|^2_F
    \end{equation}
where the hat $\hat{w}=\mathcal{F}(w)$ denotes the discrete Fourier transform (DFT) of the filter $w$, $\lambda\geq0$ is a regularization coefficient to counter model overfitting and $\scriptstyle\bigodot$ indicates Hadamard product. The solution can be quickly computed as~\cite{henriques2015}:
    \begin{equation} \label{eq:2}
        \hat{w}^k_d=\frac{\hat{x}_d^k\scriptstyle\bigodot\textstyle\hat{y}^*}{\sum_{d=1}^D\hat{x}_d^k\scriptstyle\bigodot\textstyle\hat{x}_d^{k*}+\lambda}
    \end{equation}
Here, $y^*$ represents the complex conjugate of a complex number$y$.

For the detection stage, we aim to acquire the target location in the search frame. Let $z^k\in \mathbb{R}^{V\times H\times D}$ indicate the new feature map of $k$-th CNN layer in the current frame. We transform it to the Fourier domain $\hat{z}^k=\mathcal{F}(z^k)$, and then the responses can be computed as:
    \begin{equation} \label{eq:3}
        R^k=\mathcal{F}^{-1}(\hat{z}^k\cdot\hat{w}^k)
    \end{equation}
where $\mathcal{F}^{-1}(\cdot)$ denotes the inverse discrete Fourier transform (IDFT) and $R^k\in \mathbb{R}^{V\times H}$ is the $k$-th response map resized to the same size of the image patch.

Now we obtain $K$ response maps $R=\{R^1,R^2,\ldots,R^k\}$ and our goal is to fuse all response maps into an enhanced one, denoted $Q\in \mathbb{R}^{V\times H}$. Similar to~\cite{liu2017}, we can treat the fusion as the measurement of correlation between the original response maps $R$ and the fused map $Q$. Hence, we exploit KL divergence-based method to measure this correlation and ensemble response maps. The desired fused response map $Q$ can be optimized by minimize the distance between the response maps R and the fused response map $Q$:
    \begin{equation} \label{eq:4}
        \begin{aligned}
        &\arg\min_Q\sum_{k=1}^{K}KL(R^k\|Q)\\
        &s.t. \sum q_{i,j}=1
        \end{aligned}
    \end{equation}
where $KL(R^k\|Q)=\sum_{i,j} r_{i,j}^k\log{\frac{r_{i,j}^l}{q_{i,j}}}$ denotes the KL divergence, the subscript $(i,j)$ denote the $(i,j)$-th elements of a matrix. Then, the solution can be deduced by the Lagrange multiplier method:
    \begin{equation} \label{eq:6}
        Q=\sum_{k=1}^K\frac{R^k}{K}
    \end{equation}

Finally, the target position is regarded as the location of the largest response on the fused response map $Q$.

\subsection{Re-detection Module}
\label{subsec:32}
The practical tracking environment always undergoes variations, hence the translation estimation module must be robust against challenging conditions like fast motion, illumination variation and occlusion. As Wang et al. described in~\cite{wang2017}, the detection may be disturbed by similar objects or certain noises in the search region, and this can be reflected as multiple peaks in the response map. This provides an intuitional approach to avoid tracker drift and improve localization precision by re-detecting all possible peaks exhaustively. However, examining all candidate peaks is very time consuming, and re-detecting too many similar objects may lead to inaccuracy.

Similar to~\cite{wang2017}, we propose a reliable module to ensure the efficiency and robustness of our detector. For the fusion response map $Q(z;w)$ produced by correlation, the multiple peaks are computed by
    \begin{equation} \label{eq:8}
        G(z)=Q(z;w)\cdot B
    \end{equation}
where $B$ is a binary matrix with the same size of $Q(z;w)$, whose non-zeros elements identifies the local maxima in $Q(z;w)$. Elements at the location of local maxima in $B$ are set to 1, while the other elements are set to 0.

    \begin{figure*}[t]
        \begin{center}
        \begin{minipage}{\linewidth}\centerline{\includegraphics[width=\textwidth]{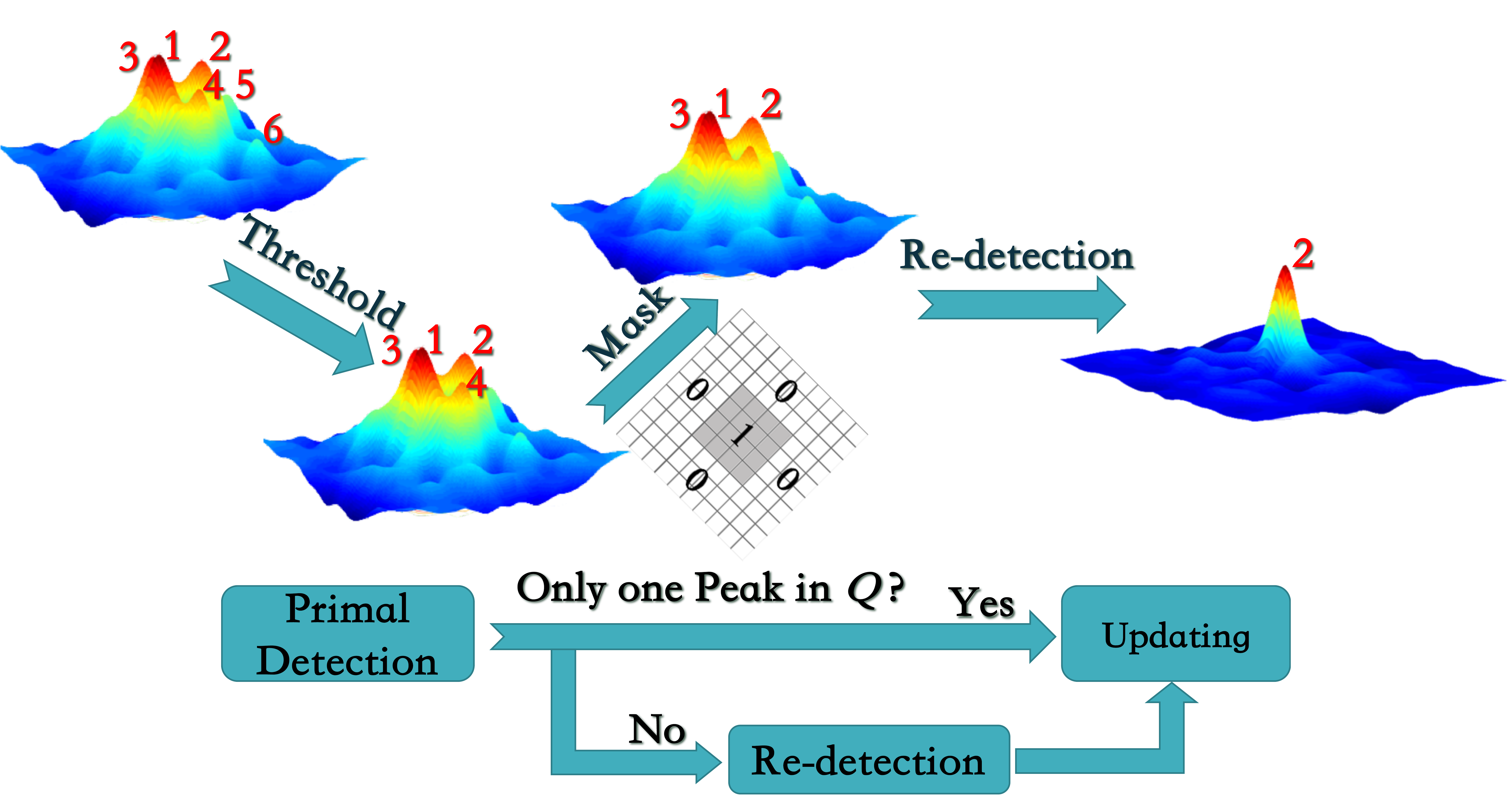}}\end{minipage}
        \end{center}
        \caption{The process of producing candidate locations for re-detecting. The numbers in each stage indicate the candidate peaks before re-detection. The target will be given by re-detection module finally.}
        \label{fig:2}
    \end{figure*}

Without loss of generality, we can always assume that the target has limited translation between consecutive frames, and that the tracker is able to detect it in a constricted search region. It reveals two practical sense that a) the object has a low probability to locate on the boundary and b) it has high chance to appear at the higher peaks. Motivated by these assumptions, we cover a mask $M$ on $B$ to constrain the candidate peaks and reformulate the function:
    \begin{equation} \label{eq:9}
        G(z)=Q(z;w)\cdot B\cdot M
    \end{equation}
where $M$ denotes a binary mask matrix with the same size of $Q(z;w)$. The elements of central region in $M$ are set to 1 (the region of ones has a proportion $\xi$ to the size of the response map), while elements in marginal region are set to 0. For each peak, a ratio between its magnitude and that of the highest peak is calculated. Those peaks whose ratio is above a pre-defined threshold $\theta$ are called as qualified peaks. Then, all qualified peaks are sorted according to its ratio, and only the top \emph{n} peaks are selected as candidate re-detection locations. The whole process is illustrated in Fig.~\ref{fig:2}.

If there existing more than one peak, the corresponding image regions centered at these candidate locations can be re-detected according to Eq.~\ref{eq:3}. Then, the location of the target can be identified as where the maximum response occurs.

\subsection{Adaptive Online Model Update}
\label{subsec:34}
The online model update is an essential step to adapt appearance variations of the target. The conventional strategy is to update linearly, which may result in tracking failure once the detection is inaccurate. Inspired by previous work, we design an adaptive online model update strategy, which aims to regulate update scheme of the model automatically.

As the maximum score of one response map indicates the degree of correlation between the object and the model learned by CF, we measure the confidence of current frame's detection result through historical scores. Assume the maximum score in current score map is $H^t$, and the historical maximum scores of previous frames are $\{H^{t-T}|T=1,\ldots,n\}$. So, we can measure the variance between current score and historical scores appropriately by:
    \begin{equation} \label{eq:12}
        C^t=S^t-\sum_{T=0}^n\frac{H^{t-T}}{n+1}
    \end{equation}

We define a basic learning rate $\eta$, which balances the proportion between the old model and the new model. The whole adaptive update strategy can be summarized as:
    \begin{equation} \label{eq:13}
        \eta^t=\left\{
        \begin{array}{lll}
        \eta        &\quad C^t>\tau\\
        0           &\quad C^t<-\tau\\
        \eta(1+C^t) &\quad \textrm{otherwise}
        \end{array}\right.
    \end{equation}
    \begin{equation} \label{eq:14}
        \hat{w}^t=(1-\eta^t)\hat{w}^{t-1}+\eta^t\hat{w}
    \end{equation}

Learning rate $\eta$ is replaced by the current adaptive learning rate $\eta^t$, Eq.~\ref{eq:14} represents that once the absolute value of variance is larger than the threshold $\tau$, which indicates the detection is inaccuracy or overconfident, the update will be shut down or take the basic learning rate. Otherwise, update process will do self-adaptive according to the current variance.

\section{Experimental Evaluations}
\label{sec:4}
Experimental evaluations are conducted on three modern tracking benchmarks: OTB-2013~\cite{wu2013}, OTB-2015~\cite{wu2015} and VOT-2017~\cite{vot2017}. All the tracking results use the raw results published by trackers own to ensure a fair comparison. We explored both quantitative and qualitative analysis with state-of-the-art trackers.

\subsection{Evaluation Setup}
We adopt the VGG-m-2048~\cite{chatfield2014} for extracting convolutional features~(Conv-1, Conv-3 and Conv-5 are employed in the experiments). And due to the lack of spatial information and to retain computation efficiency, all layers after layer 15 are removed. We crop search region with twice the size of the target. Then, we resize it to 224$\times$224 pixels to satisfy the VGG-Net input demand. The regularization coefficient is set to $\lambda=10^{-4}$ in Eq.~\ref{eq:1}. For the re-detection module, the proportion $\xi$ of the region set to 1 in mask $M$ is 0.4 and the threshold $\theta$ is set to 0.7 with the qualified top $n=3$ peaks selected as candidate re-detection locations~(refer to section~\ref{subsec:32} for more details). Similar to~\cite{danelljan2014}, we adopt patch pyramid with the scale factors $\{a^n|a=1.02,n\in ([-\frac{s-1}{2}],\ldots,[\frac{s-1}{2}])\}$ for scale estimation. Finally, for the adaptive online model update strategy, we set the threshold $\tau$ to 0.05 and the basic learning rate $\eta$ is initialed by 0.01 in Eq.~\ref{eq:14}. Our experiments are conducted in MATLAB R2015b and use the MatConvNet toolbox~\cite{vedaldi2015} on a PC with an Intel i7 3770K 3.5GHz CPU, 8G RAM, and a GeForce GTX 960 GPU in this work.

\subsection{Evaluation Metrics}
\textbf{OTB dataset:} OTB-2013~\cite{wu2013} contains 50 fully annotated sequences, and OTB-2015~\cite{wu2015} is the extension of OTB-2013 including 100 video sequences. The evaluation is based on two metrics: precision plot and success plot. The precision plot shows the percentage of frames in which the estimated locations are within a given threshold distance of the ground-truth positions. In addition, the value when threshold is 20 pixels is always taken as the representative precision score. The success plot shows the ratio of successful frames when the threshold varies from 0 to 1, where a successful frame means its overlap score is larger than the given threshold.
The rank of tracking algorithms is always given by the area under curve (AUC) of each success plot.

\textbf{VOT dataset:} VOT-2017~\cite{vot2017} contains 60 videos, three metrics are used to evaluate the performance: accuracy, robustness and average overlap. The accuracy is defined as average overlap with annotated groundtruth during successful tracking periods. The robustness is defined as how many times the trackers fail to localize the target during tracking. And the expected average overlap~(EAO) is an estimator of the average overlap a tracker is expected to attain on a large collection of short-term sequences.

\subsection{Ablation Study}
To verify our claims and justify our design choice in MLCFT, we conduct several ablation experiments. We first conduct tests with different versions of MLCFT on OTB benchmarks. We denote MLCFT without re-detection module as MLCFT-no-rd and with linear update as MLCFT-ue. OP indicates area under curve of each success plot and DP represents precision score at the threshold of 20 pixels.

    \begin{table}[t]
        \begin{center}
        \caption{Ablation Studies of MLCFT on OTB benchmarks}
        \label{table:1}
        \begin{tabular}{l|cccccc}
        \hline
        \multirow{2}{*}{} & \multicolumn{3}{c}{OTB-2013} & \multicolumn{3}{c}{OTB-2015} \\ \cline{2-7}
                                  & OP (\%)   & DP (\%)   & FPS      & OP (\%)   & DP (\%)   & FPS   \\ \hline
        MLCFT             & 67.2         & 88.4          & 16.19    & 66.4         & 87.6         & 15.86 \\
        MLCFT-nrd       & 66.2         & 87.6          & 18.93    & 63.5        & 83.2         & 19.74 \\
        MLCFT-ue         & 66.8         & 87.9          & 17.96    & 64.4        & 84.6         & 18.41 \\ \hline
        \end{tabular}
        \end{center}
    \end{table}

    \begin{figure*}[t]
        \begin{center}
        \begin{minipage}{\linewidth}\centerline{\includegraphics[width=\textwidth]{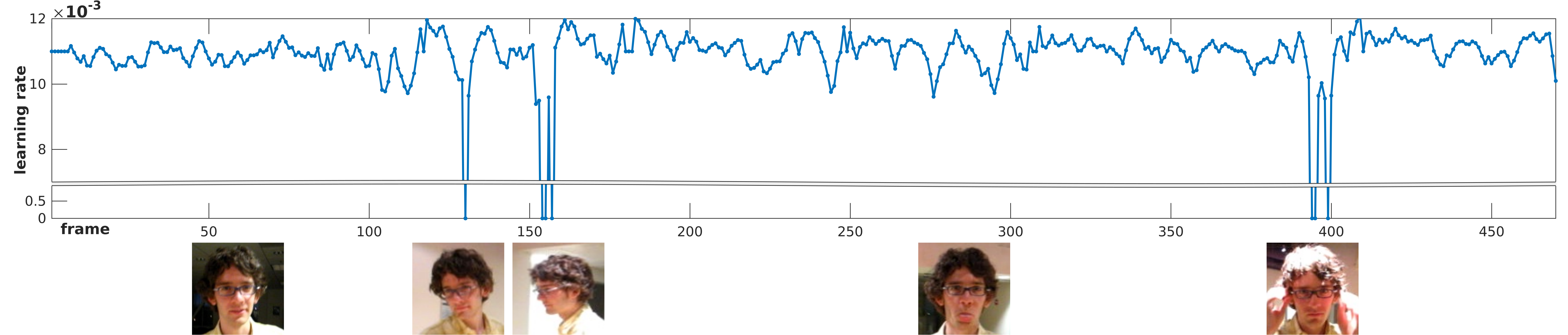}}\end{minipage}
        \end{center}
        \caption{Adaptive learning rate for the frames in the sequence \emph{David}.}
        \label{fig:3}
    \end{figure*}

As shown in Table~\ref{table:1}, MLCFT outperforms well above MLCFT-no-rd and MLCFT-ue. The results show the importance of the re-detection module, without which AUC may decrease by 1\% and 2.9\% in OTB-2013 and OTB-2015, respectively.

We visualize the variation of the adaptive learning rate during tracking to demonstrate our proposal. As shown in Fig.~\ref{fig:3}, the learning rate fluctuates in a small range, i.e., to adaptively adjust itself during the ideal conditions as frame 57 and frame 278 show. However, when the target suffers from significant deformation (frame 130), out-of-plane rotation (frame 158) and illumination variation (frame 307), the learning rate will decrease to 0. The adaptive update module automatically suspends updating process to avoid updating the model with the unconfident target.

    \begin{figure*}[t]
        \begin{center}
        \begin{minipage}[b]{0.49\linewidth}\centerline{\includegraphics[width=\textwidth]{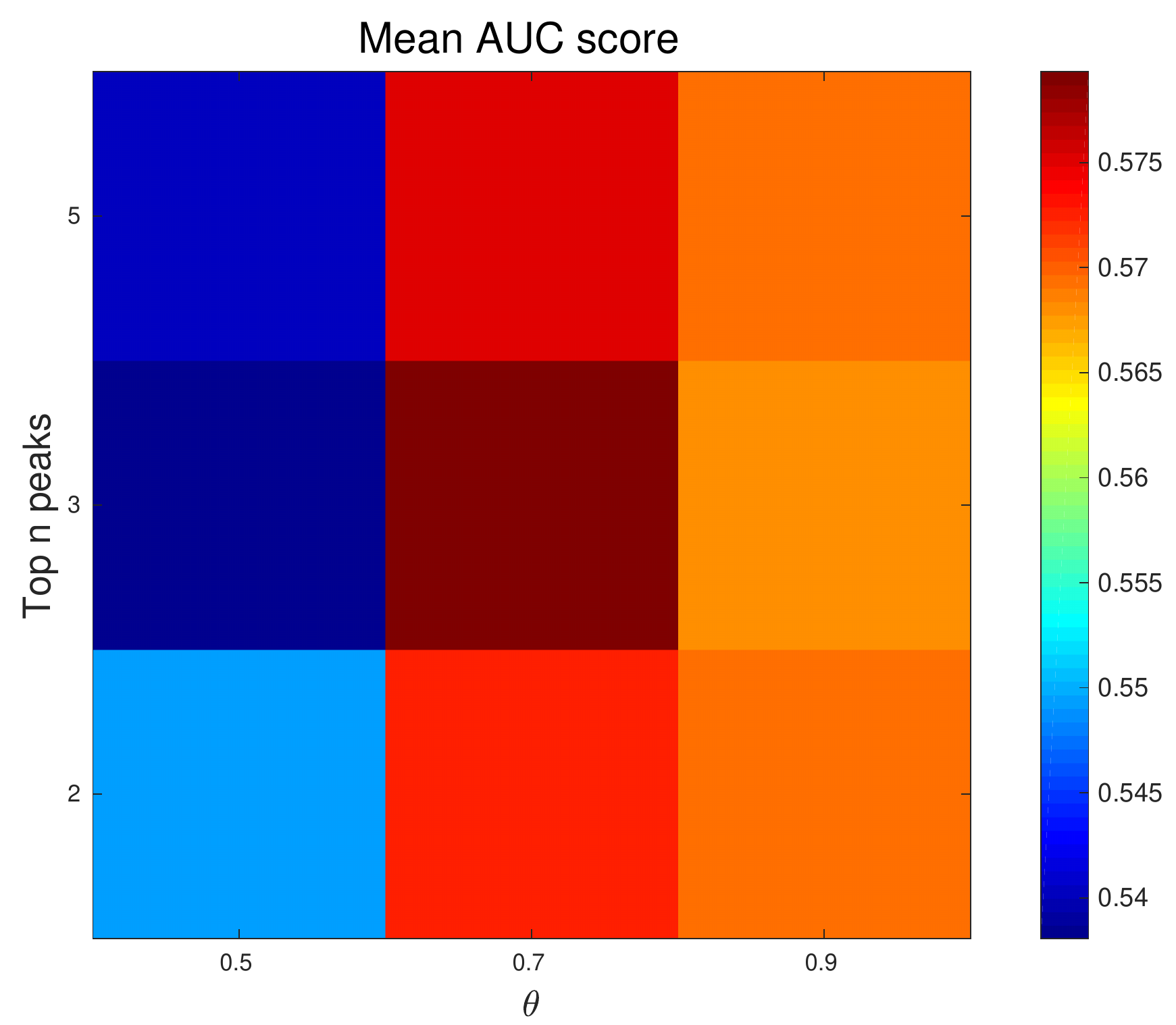}}\medskip\end{minipage}
        \begin{minipage}[b]{0.49\linewidth}\centerline{\includegraphics[width=\textwidth]{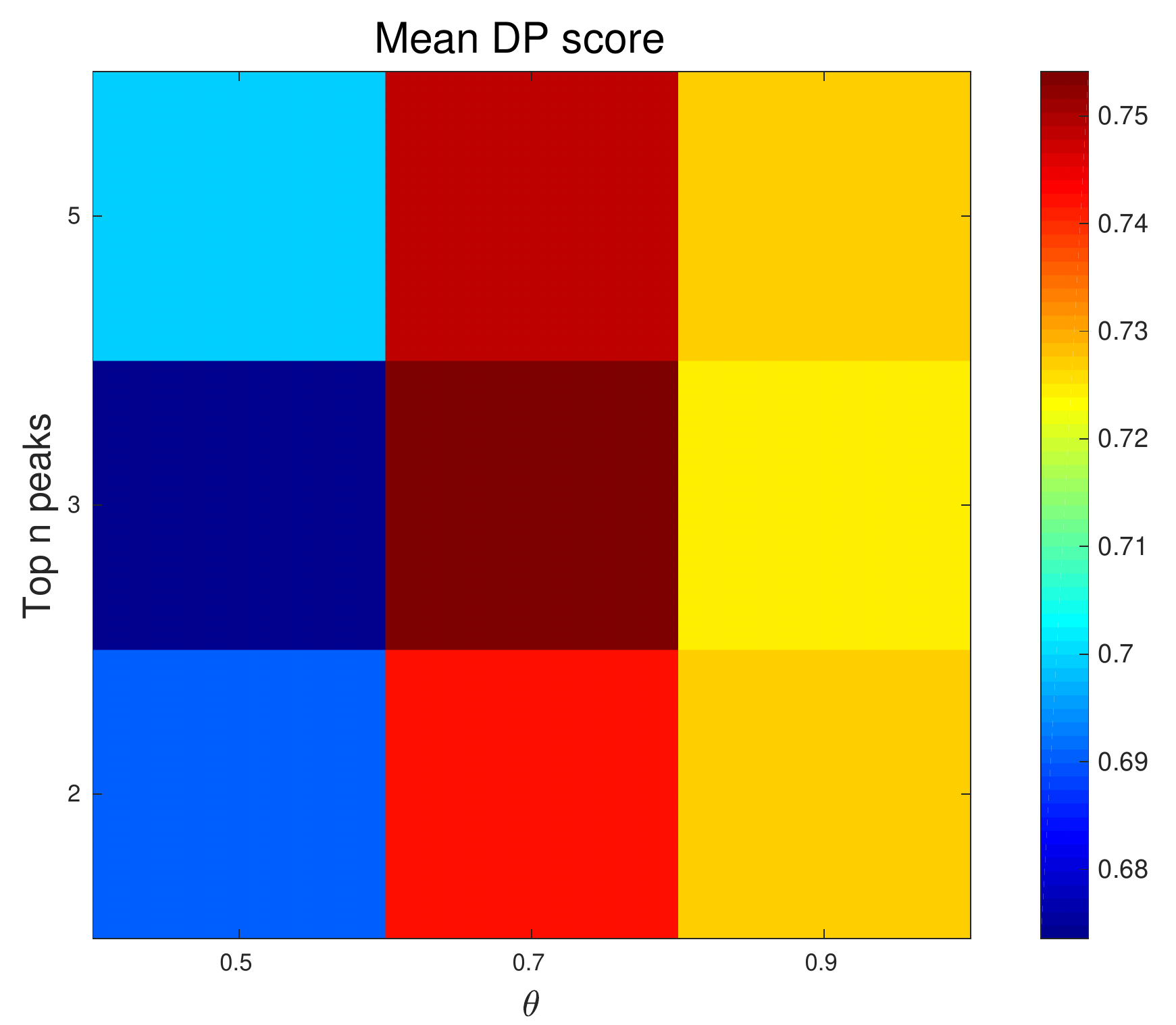}}\medskip\end{minipage}
        \end{center}
        \caption{Impact of parameters $\theta$ and top $n$ peaks on tracking results used in re-detection module on the OTB-2013 benchmark. The results are scaled in mean AUC score and mean DP score. Best viewed in color.}
        \label{fig:4}
    \end{figure*}

Moreover, in re-detection module, we set two priori thresholds as $\theta$ and $n$. In order to demonstrate the choice, the sensitivity analysis is made. As shown in Fig.~\ref{fig:4}, the best tracking result is always obtained by $\theta$=0.7 and $n$=3 both in the mean AUC score and mean DP score. And the tracker performs more sensitive to the threshold $\theta$ than the parameter $n$, i.e., the tracking result varies more dramatically in the horizontal direction than the vertical direction.

    \begin{figure*}[t]
        \begin{center}
        \begin{minipage}[b]{0.45\linewidth}\centerline{\includegraphics[width=\textwidth]{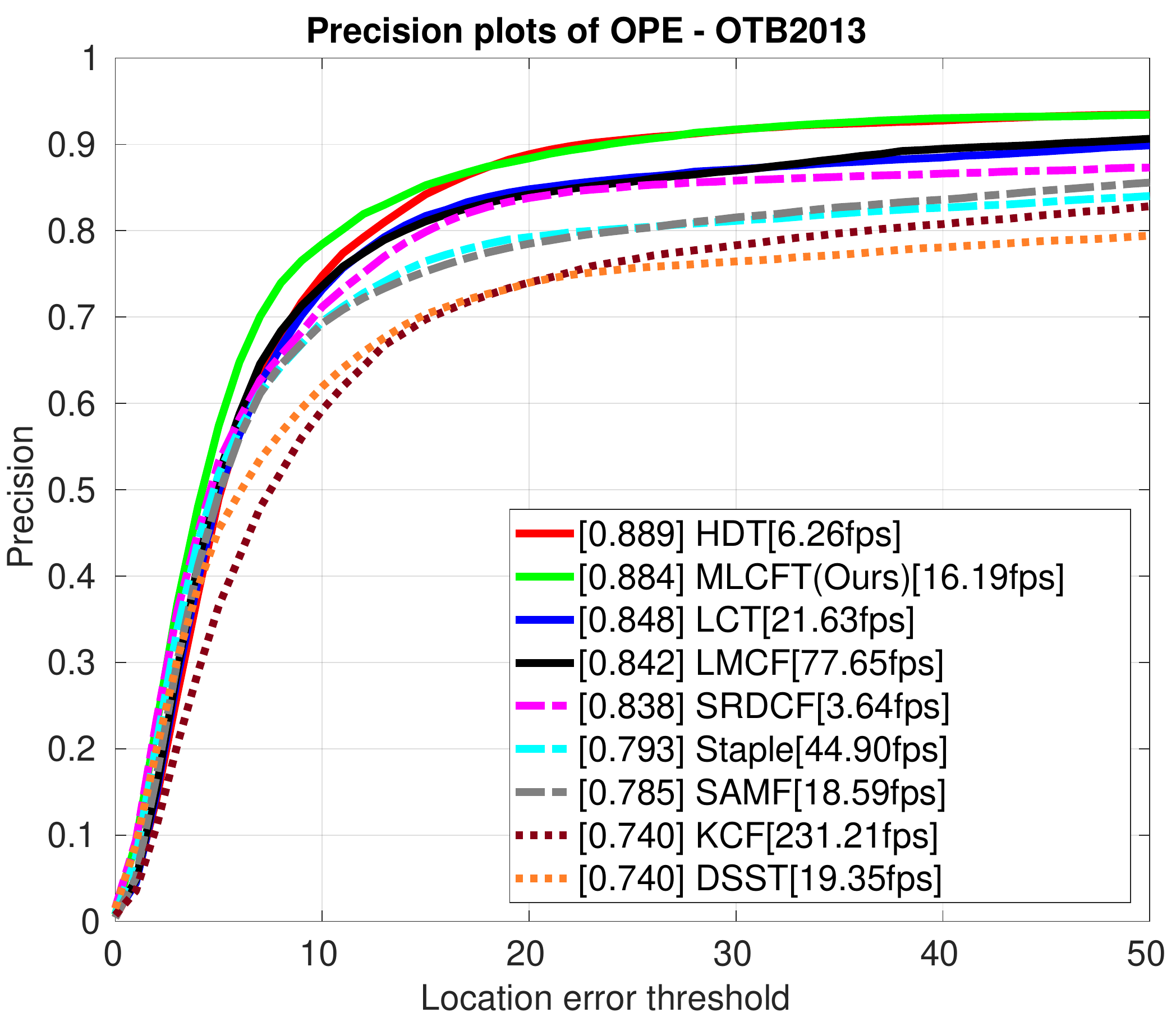}}\medskip\end{minipage}
        \begin{minipage}[b]{0.45\linewidth}\centerline{\includegraphics[width=\textwidth]{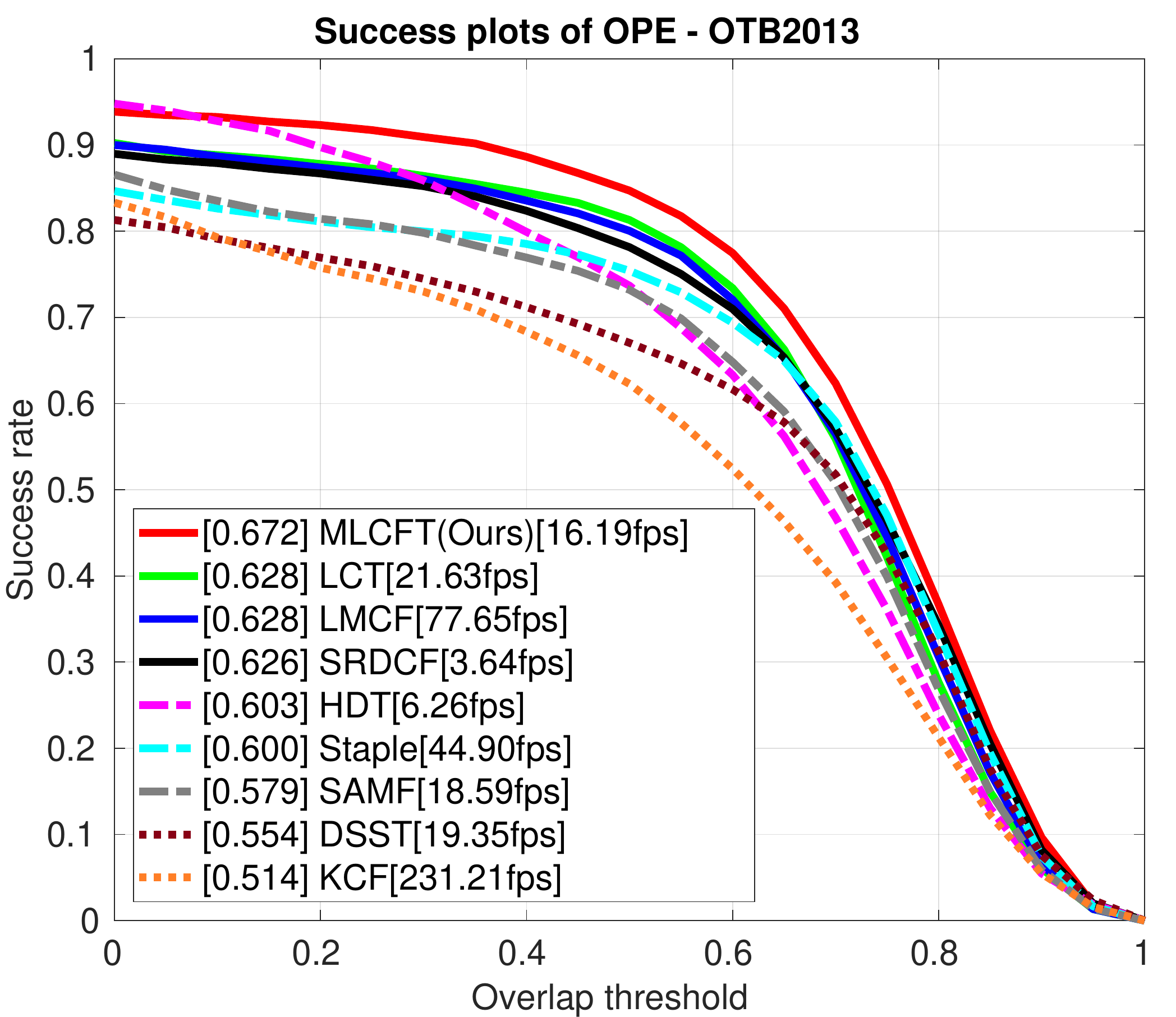}}\medskip\end{minipage}
        \end{center}
        \caption{Precision and success plots on OTB-2013. The forward numbers in the legend indicate the representative precisions at 20 pixels for precision plots, and AUC for success plots. The backward numbers in the legend denotes the speed of trackers. Best viewed in color.}
        \label{fig:5}
    \end{figure*}

\subsection{Evaluation on OTB-2013}
We evaluate our approach MLCFT with 8 state-of-the-art trackers designed with various improvement, including KCF~\cite{henriques2015}, DSST~\cite{danelljan2014}, SAMF~\cite{li2014}, SRDCF~\cite{danelljan2015a}, LMCF~\cite{wang2017}, LCT~\cite{ma2015long}, HDT~\cite{qi2016} and Staple~\cite{bertinetto2016}. The one-pass evaluation (OPE) is employed to compare these trackers.

Fig.~\ref{fig:5} illustrates the precision and success plots of compared trackers. It clearly demonstrates that our proposed tracker MLCFT outperforms those 8 compared trackers significantly in both metrics. Our approach obtains an AUC score of 0.672 in the success plot. Compared with  LMCF and LCT which are assembled with re-detection module, MLCFT gains the improvement of 4.4\%. And in the precision plot, our approach obtains a score of 0.884, outperforms LMCF and LCT by 4.2\% and 3.6\%, respectively.

\subsection{Evaluation on OTB-2015}
In this experiment, we compare our method against most recent trackers, including ECO~\cite{danelljan2017}, CREST~\cite{song2017}, BACF~\cite{galoogahi2017}, MCPF~\cite{zhang2017}, SINT~\cite{tao2016}, SRDCFdecon~\cite{danelljan2016a}, DeepLMCF~\cite{wang2017}, DeepSRDCF~\cite{danelljan2015} and KCF~\cite{henriques2015}. The OPE is also employed to compare these trackers.

\begin{figure*}[t]
        \begin{center}
        \begin{minipage}[b]{0.49\linewidth}\centerline{\includegraphics[width=\textwidth]{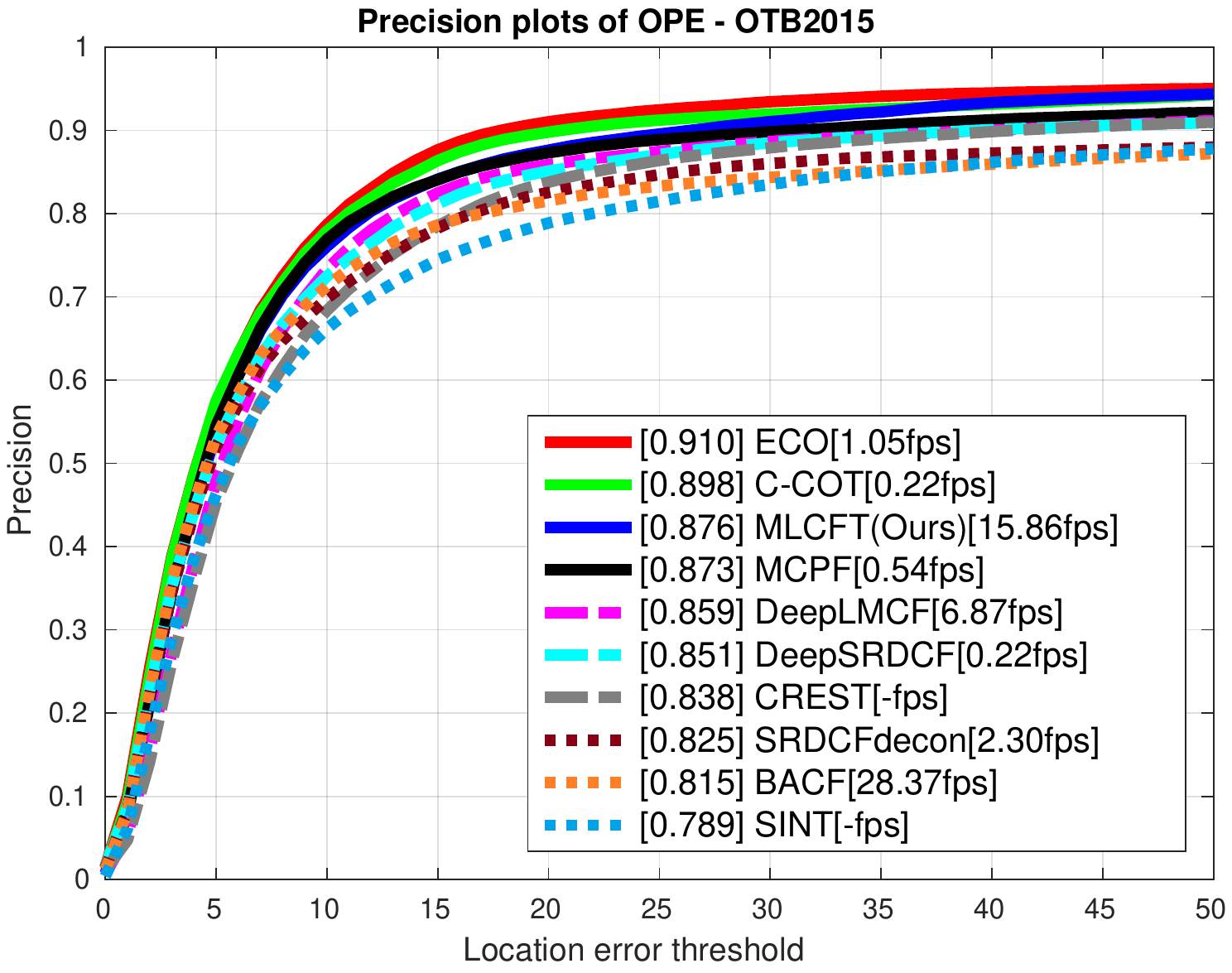}}\medskip\end{minipage}
        \begin{minipage}[b]{0.49\linewidth}\centerline{\includegraphics[width=\textwidth]{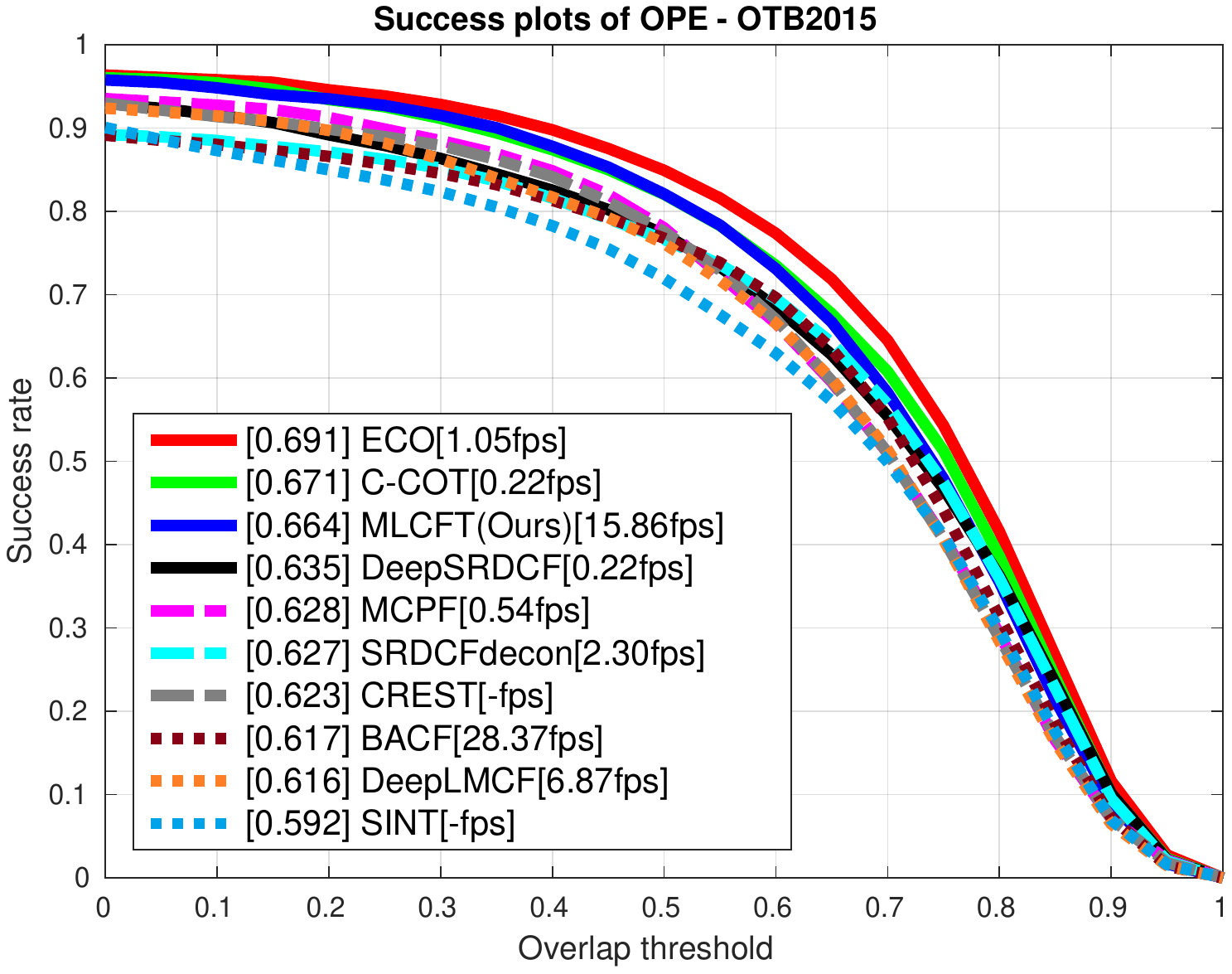}}\medskip\end{minipage}
        \end{center}
        \caption{Precision and success plots on OTB-2015. The forward numbers in the legend indicate the representative precisions at 20 pixels for precision plots, and AUC for success plots. The backward numbers in the legend denotes the speed of trackers. Best viewed in color.}
        \label{fig:6}
    \end{figure*}

The precision plots and success plots are illustrated in Fig.~\ref{fig:6}. MLCFT is close to state-of-the-art in terms of accuracy and is the fastest among all top performers. The CF-based trackers C-COT and its improved version ECO both suffer from low speed, while MLCFT does not sacrifice too much run-time performance due to the simpler framework and the cascade detection scheme based on CF. Quantitatively, compared with the top rank algorithms C-COT~\cite{danelljan2016} and the subsequent ECO~\cite{danelljan2017} tracker, MLCFT sacrifices the accuracy by 1.0\% and 3.9\% in average under curve~(AUC) but provides an 80X and 16X speedup on the OTB-2015 dataset.

The top performance can be attributed to several aspects. Firstly, our method exploits an effectively oriented re-detection module based on CF, which not only avoids the target drift but also retains the computation efficiency. Besides, the ensemble of multiple layers of convolutional layers provides expressive features to represent the target. Finally, the adaptive update strategy also contributes to the improvement of the performance and the acceleration of speed.

\subsection{Evaluation on VOT-2017}
We compare our MLCFT with state-of-the-art approaches on the VOT-2017 benchmark~\cite{vot2017}, including C-COT~\cite{danelljan2016}, CFCF~\cite{cfcf}, CFWCR~\cite{cfwcr}, CSRDCF~\cite{csrdcf} and ECO~\cite{danelljan2017}. Table~\ref{table:2} shows the comparison results over all the sequences on VOT-2017. Among these approaches, CFWCR achieves the best EAO score of 0.303 and ECO gets the best robustness score of 1.117. CFCF achieves the best accuracy score of 0.509. Meanwhile, our MLCFT obtains a second-best robustness score of 1.132 with state-of-the-art EAO score of 0.272 and accuracy score of 0.479. Our MLCFT can be regarded as state-of-the-art tracking approach since the EAO score exceeds the state-of-the-are bound which is defined as 0.251 according to the VOT-2017 report.

    \begin{table}[t]
        \small
        \begin{center}
        \caption{Comparison of MLCFT with state-of-the-art trackers on VOT-2017 benchmark. The strict state-of-the-art bound is 0.251 under EAO metrics}
         \label{table:2}
        \begin{tabular}{c|ccccccc}
        \hline
                         & C-COT & CFCF   & CFWCR & CSRDCF & ECO   & MLCFT \\ \hline
        EAO          & 0.267    & 0.286  & 0.303      & 0.256       & 0.281  & 0.272 \\
        Accuracy   & 0.493    & 0.509  & 0.484      & 0.488       & 0.483  & 0.479 \\
        Robustness & 1.315   & 1.169  & 1.210      & 1.309       & 1.117  & 1.132 \\ \hline
        \end{tabular}
        \end{center}
    \end{table}

\subsection{Qualitative Evaluation}
\begin{figure*}[t]
        \begin{center}
        \begin{minipage}[b]{\linewidth}\centerline{\includegraphics[width=\textwidth]{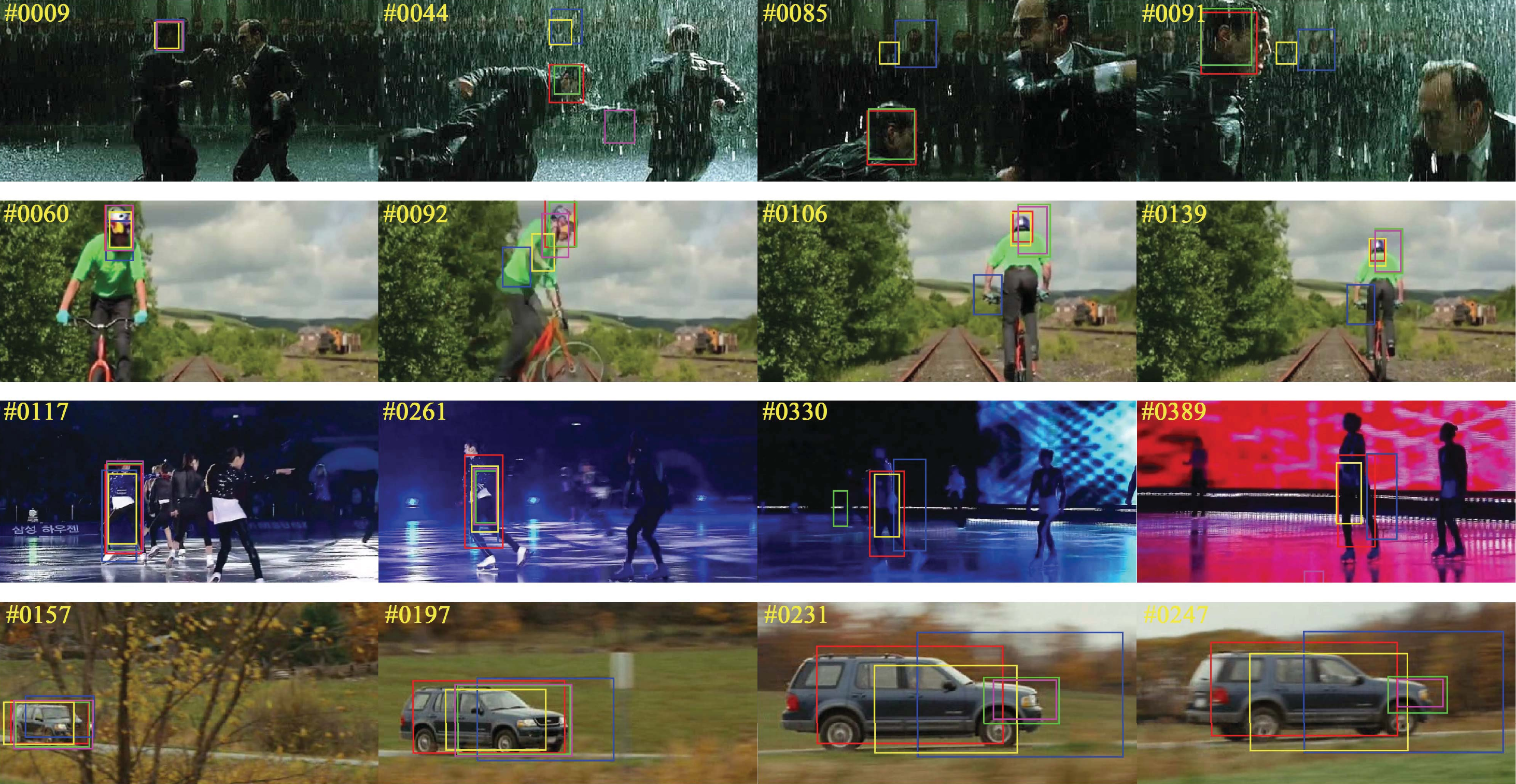}}\end{minipage}
        \vfill\begin{minipage}[b]{\linewidth}\centerline{\includegraphics[width=\textwidth]{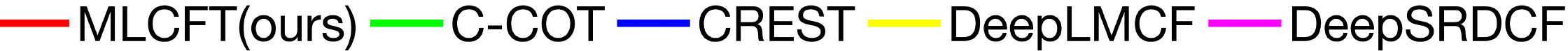}}\medskip\end{minipage}
        \end{center}
        \caption{Comparisons of our approach with state-of-the-art trackers in challenging scenarios of fast motion, significant deformation, illumination variation and scale variation on the \emph{Matrix}, \emph{Biker}, \emph{Singer2} and \emph{CarScale} sequences. Our tracker is able to avoid target drift more effectively and handle the appearance changes more accurately.}
        \label{fig:7}
    \end{figure*}

Qualitative comparisons of our approach with existing state-of-the-art trackers are conducted on OTB benchmarks. Fig.~\ref{fig:7} illustrates four challenging sequences named \emph{Matrix}, \emph{Biker}, \emph{Skater2} and \emph{CarScale} from top to bottom. In the sequence \emph{Matrix} with fast motion and background clutter, both MLCFT and C-COT can handle the translation estimation well, while MLCFT is more capable of dealing with fast scale variations.

In the sequence \emph{Biker}, the compared trackers fail to track under simultaneous fast scale variations and significant deformation. However, MLCFT and DeepLMCF accurately estimate the scale and position, as shown in frame 106. This is attributed to the re-detection module assembled in the tracker, which enhance the robustness of the tracker.

In the sequence \emph{Singer2}, all compared trackers fail or cannot handle scale variations due to the varying lighting conditions. In contrast, our approach provides better robustness and accuracy in these conditions. Additionally, in the sequence \emph{CarScale} with significant scale variations, MLCFT have the best scale estimation among all the compared trackers.

\section{Conclusions}
\label{sec:5}
In this paper, we propose a multi-level CF-based approach for visual object tracking. With an auxiliary CF-based re-detection module, our tracker shows a satisfactory robustness under challenging scenarios like fast motion, illumination variation and occlusion. Meantime, the characteristic features from multiple CNN layers offer our tracker more capability to discriminate the target. Moreover, our proposed adaptive online model update strategy can automatically handle the adjustment of the model variations, and thus promote the performance of the tracker further. Both quantitative and qualitative evaluations are performed to validate our approach. The results demonstrate that our proposed tracker obtain the state-of-the-art performance according to VOT-2017 report, while retaining higher speed than other state-of-the-art trackers.

~\\
\textbf{Acknowledgements }
This work is supported by the NSFC project under Grant No.~U1833101, Shenzhen Science and Technologies project under Grant No. JCYJ20160428182137473, the Science and Technology Planning Program of Guangdong Province under Grant No.~2016B090918047, and the Joint Research Center of Tencent \& Tsinghua.

The authors would like to thank all the anonymous reviewers for their constructive comments and suggestions.

%
%
%
\bibliographystyle{splncs04}
\bibliography{accv2018cameraready}

\begin{thebibliography}{10}
\providecommand{\url}[1]{\texttt{#1}}
\providecommand{\urlprefix}{URL }
\providecommand{\doi}[1]{https://doi.org/#1}

\bibitem{bertinetto2016}
Bertinetto, L., Valmadre, J., Henriques, J.F., Vedaldi, A., Torr, P.H.:
  Fully-convolutional siamese networks for object tracking. In: European
  Conference on Computer Vision (ECCV). pp. 850--865 (2016)

\bibitem{bolme2010}
Bolme, D.S., Beveridge, J.R., Draper, B.A., Lui, Y.M.: Visual object tracking
  using adaptive correlation filters. In: IEEE Conference on Computer Vision
  and Pattern Recognition (CVPR). pp. 2544--2550 (2010)

\bibitem{chatfield2014}
Chatfield, K., Simonyan, K., Vedaldi, A., Zisserman, A.: Return of the devil in
  the details: Delving deep into convolutional nets. arXiv preprint
  arXiv:1405.3531  (2014)

\bibitem{danelljan2017}
Danelljan, M., Bhat, G., Khan, F.S., Felsberg, M.: Eco: Efficient convolution
  operators for tracking. In: Conference on Computer Vision and Pattern
  Recognition (CVPR). pp. 21--26 (2017)

\bibitem{danelljan2014}
Danelljan, M., H{\"a}ger, G., Khan, F., Felsberg, M.: Accurate scale estimation
  for robust visual tracking. In: British Machine Vision Conference (BMVC)
  (2014)

\bibitem{danelljan2015}
Danelljan, M., Hager, G., Shahbaz~Khan, F., Felsberg, M.: Convolutional
  features for correlation filter based visual tracking. In: International
  Conference on Computer Vision Workshops (ICCVW). pp. 58--66 (2015)

\bibitem{danelljan2015a}
Danelljan, M., Hager, G., Shahbaz~Khan, F., Felsberg, M.: Learning spatially
  regularized correlation filters for visual tracking. In: International
  Conference on Computer Vision. pp. 4310--4318 (2015)

\bibitem{danelljan2016a}
Danelljan, M., Hager, G., Shahbaz~Khan, F., Felsberg, M.: Adaptive
  decontamination of the training set: A unified formulation for discriminative
  visual tracking. In: Computer Vision and Pattern Recognition. pp. 1430--1438
  (2016)

\bibitem{danelljan2016}
Danelljan, M., Robinson, A., Khan, F.S., Felsberg, M.: Beyond correlation
  filters: Learning continuous convolution operators for visual tracking. In:
  European Conference on Computer Vision. pp. 472--488 (2016)

\bibitem{galoogahi2017}
Galoogahi, H.K., Fagg, A., Lucey, S.: Learning background-aware correlation
  filters for visual tracking. In: IEEE Conference on Computer Vision and
  Pattern Recognition (CVPR). pp. 21--26 (2017)

\bibitem{pgao2018ieice}
Gao, P., Ma, Y., Li, C., Song, K., Zhang, Y., Wang, F., Xiao, L.: Adaptive
  object tracking with complementary models. IEICE Transactions on Information
  and Systems  \textbf{E101.D}(11) (2018)

\bibitem{pgao180407459}
Gao, P., Ma, Y., Song, K., Li, C., Wang, F., Xiao, L.: A complementary tracking
  model with multiple features. arXiv preprint arXiv:1804.07459  (2018)

\bibitem{pgao2018}
Gao, P., Ma, Y., Song, K., Li, C., Wang, F., Xiao, L.: Large margin structured
  convolution operator for thermal infrared object tracking. In: International
  Conference on Pattern Recognition (ICPR). pp. 2380--2385 (2018)

\bibitem{pgao180408208}
Gao, P., Ma, Y., Song, K., Li, C., Wang, F., Xiao, L., Zhang, Y.: High
  performance visual tracking with circular and structural operators.
  Knowledge-Based Systems  (2018)

\bibitem{girshick2014}
Girshick, R., Donahue, J., Darrell, T., Malik, J.: Rich feature hierarchies for
  accurate object detection and semantic segmentation. In: IEEE Conference on
  Computer Vision and Pattern Recognition (CVPR). pp. 580--587 (2014)

\bibitem{cfcf}
Gundogdu, E., Alatan, A.: Good features to correlate for visual tracking. arXiv
  preprint arXiv:1704.06326  (2017)

\bibitem{cfwcr}
He, Z., Fan, Y., Zhuang, J., Dong, Y., Bai, H.: Correlation filters with
  weighted convolution responses. In: IEEE International Conference on Computer
  Vision (ICCV). IEEE (2017)

\bibitem{henriques2012}
Henriques, J.F., Caseiro, R., Martins, P., Batista, J.: Exploiting the
  circulant structure of tracking-by-detection with kernels. In: European
  Conference on Computer Vision (ECCV). pp. 702--715 (2012)

\bibitem{henriques2015}
Henriques, J.F., Caseiro, R., Martins, P., Batista, J.: High-speed tracking
  with kernelized correlation filters. IEEE Transactions on Pattern Analysis
  and Machine Intelligence  \textbf{37}(3),  583--596 (2015)

\bibitem{karpathy2014}
Karpathy, A., Toderici, G., Shetty, S., Leung, T., Sukthankar, R., Fei-Fei, L.:
  Large-scale video classification with convolutional neural networks. In: IEEE
  Conference on Computer Vision and Pattern Recognition (CVPR). pp. 1725--1732
  (2014)

\bibitem{vot2017}
Kristan, M., Leonardis, A., Matas, J., Felsberg, M., Pflugfelder, R.,
  \v{C}ehovin, L., Voj{\'\i}{\v{r}}, T., et~al: The visual object tracking
  vot2017 challenge results. In: International Conference on Computer Vision
  (ICCV). pp. 1949--1972 (2016)

\bibitem{lecun1998}
LeCun, Y., Bottou, L., Bengio, Y., Haffner, P.: Gradient-based learning applied
  to document recognition. Proceedings of the IEEE  \textbf{86}(11),
  2278--2324 (1998)

\bibitem{li2014}
Li, Y., Zhu, J.: A scale adaptive kernel correlation filter tracker with
  feature integration. In: European Conference on Computer Vision (ECCV). pp.
  254--265 (2014)

\bibitem{liu2017}
Liu, Q., Lu, X., He, Z., Zhang, C., Chen, W.S.: Deep convolutional neural
  networks for thermal infrared object tracking. Knowledge-Based Systems
  \textbf{134},  189--198 (2017)

\bibitem{csrdcf}
Luke{\v{z}}i{\v{c}}, A., Voj{\'\i}{\v{r}}, T., {\v{C}}ehovin, L., Matas, J.,
  Kristan, M.: Discriminative correlation filter with channel and spatial
  reliability. In: IEEE Conference on Computer Vision and Pattern Recognition
  (CVPR). pp. 4847--4856. IEEE (2017)

\bibitem{ma2015long}
Ma, C., Yang, X., Zhang, C., Yang, M.H.: Long-term correlation tracking. In:
  Proceedings of the IEEE conference on computer vision and pattern
  recognition. pp. 5388--5396 (2015)

\bibitem{qi2016}
Qi, Y., Zhang, S., Qin, L., Yao, H., Huang, Q., Lim, J., Yang, M.H.: Hedged
  deep tracking. In: IEEE Conference on Computer Vision and Pattern Recognition
  (CVPR). pp. 4303--4311 (2016)

\bibitem{sermanet2013}
Sermanet, P., Eigen, D., Zhang, X., Mathieu, M., Fergus, R., LeCun, Y.:
  Overfeat: Integrated recognition, localization and detection using
  convolutional networks. arXiv preprint arXiv:1312.6229  (2013)

\bibitem{song2017}
Song, Y., Ma, C., Gong, L., Zhang, J., Lau, R.W., Yang, M.H.: Crest:
  Convolutional residual learning for visual tracking. In: International
  Conference on Computer Vision (ICCV). pp. 2574--2583 (2017)

\bibitem{tao2016}
Tao, R., Gavves, E., Smeulders, A.W.: Siamese instance search for tracking. In:
  IEEE Conference on Computer Vision and Pattern Recognition (CVPR). pp.
  1420--1429 (2016)

\bibitem{valmadre2017}
Valmadre, J., Bertinetto, L., Henriques, J., Vedaldi, A., Torr, P.H.:
  End-to-end representation learning for correlation filter based tracking. In:
  IEEE Conference on Computer Vision and Pattern Recognition (CVPR). pp.
  5000--5008 (2017)

\bibitem{vedaldi2015}
Vedaldi, A., Lenc, K.: Matconvnet: Convolutional neural networks for matlab.
  In: ACM International Conference on Multimedia (ACMMM). pp. 689--692 (2015)

\bibitem{wang2017}
Wang, M., Liu, Y., Huang, Z.: Large margin object tracking with circulant
  feature maps. In: IEEE Conference on Computer Vision and Pattern Recognition
  (CVPR). pp. 21--26 (2017)

\bibitem{wu2013}
Wu, Y., Lim, J., Yang, M.H.: Online object tracking: A benchmark. In: Computer
  vision and pattern recognition (CVPR). pp. 2411--2418 (2013)

\bibitem{wu2015}
Wu, Y., Lim, J., Yang, M.H.: Object tracking benchmark. IEEE Transactions on
  Pattern Analysis and Machine Intelligence  \textbf{37}(9),  1834--1848 (2015)

\bibitem{zhang2017}
Zhang, T., Xu, C., Yang, M.H.: Multi-task correlation particle filter for
  robust object tracking. In: IEEE Conference on Computer Vision and Pattern
  Recognition (CVPR) (2017)

\end{thebibliography}

\end{document}